\documentclass[11pt,a4paper]{article}
\usepackage[hyperref]{ranlp2025}
\usepackage{times}
\usepackage{latexsym}

\usepackage{microtype}

\aclfinalcopy 

\usepackage{paralist}

\newcommand{\textnl}{\textsl}
\newcommand{\gloss}[1]{`\textnl{#1}'}
\newcommand{\heb}[1]{}
\newcommand{\hebgloss}[2]{\gloss{#2}}
\newcommand{\fval}[2]{#1=`#2'}

\title{An Annotation Scheme for Factuality \\and its Application to Parliamentary Proceedings}

\author{Gili Goldin  \\
Dept.\ of Comp.\ Sci.\\
  University of Haifa \\
 \texttt{gili.sommer@gmail.com} 
  \\\And
  Shira Wigderson \\
Dept.\ of Comp.\ Sci.\\
  University of Haifa\\
  \\\And
  Ella Rabinovich  \\
  The Academic College\\ of Tel-Aviv Yaffo \\
  \\\And
  Shuly Wintner  \\
Dept.\ of Comp.\ Sci.\\
  University of Haifa \\
  }

\date{}

\begin{document}
\maketitle
\begin{abstract}
Factuality assesses the
extent to which a language utterance relates to real-world information; it determines whether utterances
correspond to facts, possibilities, or imaginary situations, and as such, it is instrumental for fact checking.  
Factuality is a complex notion that relies on multiple linguistic signals, and has been studied in various disciplines. 
We present a complex, multi-faceted annotation scheme of factuality that combines concepts from a variety of previous works. We developed the scheme for Hebrew, but we trust that it can be adapted to other languages. We also present a set of almost~5,000 sentences in the domain of parliamentary discourse that we manually annotated according to this scheme. We report on inter-annotator agreement, and experiment with various approaches to automatically predict (some features of) the scheme, in order to extend the annotation to a large corpus.
\end{abstract}

\section{Introduction}

With the abundance of information and the rise of generative AI, ``fake'' information, such as fake news or fake product reviews, is becoming increasingly ubiquitous.  To evaluate the veracity of information, it is necessary to first identify which utterances are candidates for such verification.  
\emph{Factuality} assesses the
extent to which a language utterance relates to real-world information; it is
a measure that determines whether utterances
correspond to facts, possibilities, or imaginary situations.  
Factuality is a complex notion that has been studied in various disciplines, using varying domain-specific definitions and terminologies. 
The degree of
factuality with which a speaker makes a claim amalgamates values for
agency, ambiguity, authoritativeness, 
certainty, credibility,
commitment, confidence, hedging, approximation,
modality, perspective, stance, polarity, and  more.

It is important not to confuse factuality with 
\emph{veracity} or \emph{truthfulness}.  The
factuality of a proposition does not align it with ground truth facts.
Nevertheless, determining the factuality of sentences is a necessary
step toward achieving this latter goal; specifically, factuality can help
assess if a claim involves information that is potentially
\emph{fact-checkable} or \emph{check-worthy}.  

We describe a complex, multi-faceted annotation scheme of factuality that we apply to Hebrew texts (\S\ref{sec:schema}). The scheme amalgamates various linguistic and extra-linguistic cues that help identify factuality. Our ultimate goal is to annotate a sizable corpus of Hebrew parliamentary proceedings according to this scheme, thereby providing the infrastructure necessary for identifying fake information in Hebrew texts.

We manually annotated 4,987 Hebrew sentences from a corpus of parliamentary proceedings according to this scheme and assessed inter-coder agreement on this complex (and often subjective) task (\S\ref{sec:annotator_agreement}).
Next, we focused on one aspect of the scheme, namely the \emph{check-worthiness} feature, and evaluated various models on predicting this feature (\S\ref{sec:experiments}).
We show that off-the-shelf SOTA GPT models perform rather poorly on this classification task, whereas fine-tuned Hebrew LLMs that use the annotated data are much more accurate. We use the best performing model to automatically annotate the entire parliamentary corpus for this feature.%
\footnote{All the resources and code we developed are available at our \href {https://github.com/HaifaCLG/Factuality}{GitHub repository} and are released under the \href{https://creativecommons.org/licenses/by-sa/4.0/deed.en}{Creative Commons Attribution-ShareAlike 4.0 International License}.}

\section{Related Work}
\label{sec:lit-review}

Much of the information pertaining to factuality is encoded
linguistically, and various computational works employ linguistic information to
identify related patterns in text.  
To train classifiers that can predict (aspects of) factuality, annotated corpora are required, and several corpora include annotations 
that highlight information pertaining to factuality. 
Existing annotation schemes vary with respect to the basic unit for annotation (claim, sentence, paragraph, text, etc.) and the target tag (what exactly is being coded). Classifiers that identify factuality again vary with respect to the basic unit for classification, the features used for representing each instance, the actual prediction and the classification model. We survey several examples below.

\paragraph{Annotation schemes} were developed for research findings, hypotheses, and evidence-providing in scientific
articles \citep{teufel2000argumentative,shatkay2008multi}. For
example, 
%
FactBank \citep{sauri:2009} is a corpus of newswire texts with
annotations of perspective, polarity and factuality on a graded
(12-point) scale. This annotation is text-based, ``avoiding any
judgment based on knowledge of how things are in the world''
\citep[p.~137]{sauri:2008}.  FactBank was extended with subjective
judgments: 
 annotators indicated whether they believed the event described did (or will) happen \citep{deMarneffe:etal:2012}.
Based on their annotated version of FactBank,
\citet{deMarneffe:etal:2012} developed a classifier to
predict what they call \emph{veridicality},  a gradual property. 
The features used
for classification were mostly linguistic, with some external knowledge.

\citet{sauri:2012} proposed a linguistically-motivated computational
model  to distinguish facts from negated facts, qualifying them by certainty to label events as “possibly factual” or “possibly counterfactual”.
They identified lexical, morphosyntactic, and frame semantic markers relevant to two aspects of factuality: polarity and modality.
%
They refer to an event as the atomic unit for factuality, and use~19 features for each \emph{word} rather than for each sentence, as a single sentence may include more than one event. 
The features included linguistic cues, source or speaker information and event properties.
These features ultimately generate a degree of factuality, which touches on both \emph{polarity} and epistemic \emph{modality} distinctions as encoded in factuality markers, and includes also the \emph{source} assigning the factuality value to an event.
\emph{Modality} has three values: certain (CT), probable (PR), and possible (PS); polarity values are either positive ($+$) or negative ($-$). This leads to a total of~six factuality values: CT+, PR+, PS+, CT-, PR-, PS-, plus a seventh value, Uu, for underspecified.

MAVEN-Fact \citep{li2024mavenfactlargescaleeventfactuality} is a large-scale event factuality detection dataset. It includes factuality annotation of over 112K events, each labeled as one of~5 classes, based on the polarity and
modality of an event \citep{sauri:2009, ijcai2018p597}.   
All these datasets are in English, although 
datasets in other languages begin to emerge \citep[e.g.][]{DBLP:conf/clef/AtanasovaMBESZK18,DBLP:conf/clef/Barron-CedenoES18,DBLP:conf/clef/HasanainHSAHEBM20,nakov-etal-2021-covid}.

The scheme that we describe in \S\ref{sec:schema} combines various facets of factuality that are drawn from all these works and more. We focus on linguistic markers that can be identified in the text, and adapt existing approaches to the special case of Hebrew, a language with complex morphology and deficient orthography \citep{hebrew-lr,semitic-introduction,semitic-morphology}.

\paragraph{Automatic prediction of factuality} is a well-established computational task.
\emph{ClaimBuster} \citep{Hassan:2017:TAF:3097983.3098131}  classified claims in US presidential debates as
\emph{non-factual}, \emph{unimportant factual}, and \emph{check-worthy factual}.  
\citet{gencheva-etal-2017-context} evaluated factuality for full sentences, but also considered 
longer texts.
They adopted ClaimBuster's sentence level features, but modified them 
in various ways.
They also used context level features: the \emph{position} of the sentence in the segment, \emph{segment size} including the size of the previous and next segments, and \emph{metadata}. 

\citet{10.1145/3412869}  annotated sentences into~7 categories: personal experience, quantity in the past or present, correlation or causation, current laws or rules of operation, prediction, other type of claim, or not a claim. 
Each claim was also given a binary value, whether it is \emph{checkable} or not.


More recent approaches to detection of factuality mainly focus on the outputs of large language models (LLMs), in particular on identifying hallucinations; see \citet{wang-etal-2024-factuality} for a review. 
Contemporary works use LLMs to predict factuality in various domains \citep{zhang-etal-2024-transferable,azizov-etal-2024-safari,li-etal-2024-maven}, as we do here (\S\ref{sec:experiments}).

\section{Data}
\label{data}

Factuality is a core factor in all types of communication, but it is
paramount in argumentative genres like political discourse, where implicit strategies may exist
that help the authors avoid commitment, feign authoritativeness,
or keep (checkable) facts vague.  
%
Therefore, we focus in this work on the Knesset Corpus \citep{Goldin2024TheKC}, a large-scale dataset of parliamentary proceedings, including both plenum sessions and committee deliberations, spanning several decades. This dataset is also annotated with rich meta-information, linguistic features and named entities. 

We manually annotated almost~5,000 sentences from this corpus for the full factuality scheme proposed in \S\ref{sec:schema}. The sentences were sampled from protocols of both plenary sessions and committee deliberations, balancing across various factors such as year, author's gender, political affiliation, native language, and more. This manually annotated subset is available at \href {https://github.com/HaifaCLG/Factuality}{our GitHub repository}.

\section{An Annotation Scheme for Factuality}
\label{sec:schema}
We present a complex, multi-faceted annotation scheme that characterizes various aspects that contribute to factuality. 
The scheme includes  several features that can assist evaluators to verify the truth value of a proposition.
The basic unit of annotation is a sentence, but sentences that include several claims can be annotated with sequences of features, one set per claim. 
We list and exemplify various elements of the scheme and the values they take (for convenience, the original sentences are translated to English in the examples). 
More examples are given in our \href {https://github.com/HaifaCLG/Factuality}{GitHub repository}.
The scheme is  organized by \emph{layers}, each including several features.

%

\subsection{Check-worthiness}
\label{sec:check-worthiness}
The first layer of the annotation consists of a check-worthiness estimation for each claim in the sentence.
We provide several types of scores, based on previous works (see \S\ref{sec:lit-review}).

\paragraph{Check-worthiness score}
Following \citet{Hassan:2017:TAF:3097983.3098131}
we include an overall score of ``check-worthiness''. This feature can take the following possible values: \emph{worth checking}, \emph{not worth checking}, \emph{not a factual proposition}.
This score also subsumes the binary score used by \citet{gencheva-etal-2017-context}.
Typical claims that are marked `not a factual proposition' include questions and speech acts.

\paragraph{Claim type} Following \citet{10.1145/3412869}, each claim in the sentence is assigned a claim-type. The possible values are: \emph{personal experience}, \emph{quantity in the past or present}, \emph{correlation or causation}, \emph{current laws or rules of operation}, \emph{prediction}, \emph{other type of claims}, \emph{not a claim}, and \emph{irrealis mood}.
Appendix~\ref{app:konst} provides a detailed definition of these types.

\paragraph{Factuality profile} Following \citet{sauri:2012}, we assign each claim a \emph{factuality profile}: a pair that consists of the source and the factuality value that combines modality and polarity (see \S\ref{sec:lit-review}). 

\paragraph{Examples}
\begin{compactenum}
    \item \hebgloss{צר לי, אני לא מאמין.}{I'm sorry, I don't believe it.}
    \\check worthiness score = not worth checking
    \\claim type = other type of claim
    \\factuality profile = (speaker (from metadata), CT-)
    \item \hebgloss{000,007 גמלאים יש במדינת ישראל.}{There are 700,000 pensioners in the State of Israel.}
    \fval{check worthiness score}{worth checking}; \fval{claim type}{quantity in the past or present}; 
    \fval{factuality profile}{$\langle$speaker, CT+$\rangle$}.
\end{compactenum}

\subsection{Event Selecting Predicates}
\citet{sauri:2009} propose that in many cases, the factuality of events is conveyed by \emph{event selecting predicates (ESPs)}, which are predicates that select for an argument denoting an event. ESPs qualify the degree of factuality of their embedded event, which can be presented as a fact in the world, a counterfact, or a possibility.
They distinguish between two kinds of ESPs: \emph{source introducing predicates (SIPs)} (e.g., \hebgloss{אמר}{say}, \hebgloss{ידע}{know}) and \emph{non-source introducing predicates (NSIPs)} (e.g., \hebgloss{רצה}{want}). 
If an ESP is found in the sentence, we classify it as SIP or NSIP, based on whether or not the predicate introduces a new source. For SIPs we also mark the source they introduce, if present in the sentence, and mark a connection between the two. 
\paragraph{ESP} All ESPs present in the sentence are  marked. 
\paragraph{ESP type} For each ESP, is it SIP or NSIP. 

\paragraph{Examples}
\begin{compactenum}
    \item \hebgloss{אני לא זוכר שאי-פעם פנו אלי כחבר ועדת הכספים ולא נעניתי לדרישתם.}{I don't remember that they ever approached me as a member of the finance committee and I didn't respond to their request.}
    \fval{ESP}{\hebgloss{זוכר}{remember}}; 
    \fval{ESP type}{SIP};
    \fval{ESP source}{\hebgloss{אני}{I}}.
\end{compactenum}

\subsection{Agency}
The existence of an agent and its characteristics impact the factuality of a proposition. The absence of an agent blurs the meaning of the sentence, and agent-less utterances are often used to avoid commitment. Even if an agent does exist, its referent may be vague (e.g., \textnl{we}).
This component specifies the agent of the proposition, its position in the proposition, its animacy and its morphological properties, as well as the predicate it is an agent of and the relation between them.
%
\paragraph{Agent}
The (tokens making up the) agent in the proposition. If it is a pronoun, we indicate the referent, if known, separated by comma. 
\paragraph{Experiencer}
The (tokens making up the) experiencer in the proposition. If it is a pronoun, we indicate the referent, if known. 
Unlike the agent, the experiencer is not performing an action, but rather receives sensory or emotional input. 
\paragraph{Agent-less} If there is no agent in the proposition, we  explain why; values of this feature include \emph{passive w/o by-clause}, \emph{middle\_verb},
\emph{infinitival clause}, \emph{\hebgloss{יש/אין}{there is (not)} + infinitive}, \emph{impersonal modal verb}, \emph{imperative}, \emph{nominal clause}, \emph{existential} and \emph{unspecified} (we use `unspecified' when the agent exists, but has no identifiable referent).

\paragraph{Position}
    \emph{subject, indirect object,             embedded pronoun subject} or \emph{   embedded pronoun non subject}.
\paragraph{Animacy} 
    \emph{human, animate, inanimate}.
\paragraph{Morphology}
    \emph{1/2/3, sg/pl}.

\paragraph{Examples}
\begin{compactenum}
    \item\hebgloss{משרד התחבורה מקדם רפורמות נוספות, שכוללות בין היתר רפורמה בתעריפים הבין-עירוניים.}
    {The Ministry of Transportation is promoting additional reforms, which include inter-city fare reform.}
    \fval{agency agent}{\hebgloss{משרד התחבורה}
    {The Ministry of Transportation}}, 
    \fval{position}{\emph{subject}}, \fval{animacy}{\emph{inanimate}}, 
    \fval{morphology}{\emph{3sg}}, 
    \fval{agency-predicate}{\hebgloss{מקדם}{promotes}}. 
\end{compactenum}

\subsection{Stance}
This component characterizes the speaker's belief towards the proposition: the extent to which the speaker is certain or uncertain about what they claim. Does the speaker express a wish or a fact? Do they provide a reference (statistics, numbers, citations, etc.) to substantiate their proposition? 

Previous works suggested various ways to annotate stance: \citet{marin2011effective} used the distinction between effective and epistemic stance, \citet{pyatkin-etal-2021-possible} suggested a modal classification that maps into similar categories of stance, which they called \emph{priority and plausibility modality}, and \citet{sauri:2012} and \citet{deMarneffe:etal:2012} characterized stance as a double-axis scale: polarity(binary) and modality(continuous). The combination of these two characteristics constitutes the factuality profile that is part of our check-worthiness score (see \S\ref{sec:check-worthiness}).

We combine several of these suggestions to describe stance.  
Our stance marking includes an overall confidence level, whether it is effective or epistemic, the polarity, and a list of the lexical items that imply these values. 
%
%
%
\paragraph{Confidence level} The overall confidence level of the speaker towards their proposition, as evaluated by the annotator; the level of confidence is determined by the lexicon the speaker uses. 
The lexicon plays an important role in determining the confidence level of the speaker towards their statement, which is why we include a list of lexical items, mapped to each level of confidence. 
The list of lexical items is composed also of items from various sources \citep{modals, OnAssertivePredicates,gencheva-etal-2017-context}.
The values of this feature are \emph{high, mid}, \emph{low} and \emph{irrelevant} (if the sentence does not contain a proposition).

\paragraph{Stance type} We adopt the terminology of \citet{marin2011effective} and distinguish between \emph{effective} and \emph{epistemic} stance. When the speaker \textbf{explicitly} expresses \emph{their attitude} towards the proposition it is marked as an effective stance; when the speaker expresses \emph{knowledge} or \emph{estimation} regarding the proposition and the possibility of its realization it will be marked as epistemic stance. When there is no proposition in the sentence, we mark it as `irrelevant'.
Some examples of {effective} stance markers include:
\hebgloss{חייב/מוכרח}{must}, \hebgloss{בלתי אפשר}{impossible}, and \hebgloss{קיווה}{hoped}.
Examples of epistemic stance compatible items include
\hebgloss{אני בטוח}{I'm certain}, \hebgloss{עובדה היא}{the fact is}, and \hebgloss{זה ברור}{it is evident}.

\paragraph{Polarity} Following \citet{sauri:2012}, we mark the polarity of the proposition. 
The values are \emph{positive, negative}, or \emph{underspecified}.
We also indicate the lexical item(s) that helped the annotator decide the polarity of the proposition, according to a list we provide (available in the supplementary materials). 
The type and the lexical item are presented as pairs.
Note that usually, for a sentence to be positive it does not necessarily need to have a positive lexical item, it just needs to have no negative lexical items.

Negative polarity lexical items include
\hebgloss{לא}{no/not},
\hebgloss{אף פעם}{never}, \hebgloss{אף אחד}{no one}; 
positive ones include
\hebgloss{יש}{there is} and \hebgloss{אכן}{indeed}.
Underspecified is assigned when the source is not committed to the polarity of their statement. 

\paragraph{Reference} Whether the speaker mentions a source, and specifies the source's type and name (which can help determine how reliable that source is or how confident they are towards their proposition being true). 
Values are pairs of [\emph{name}, \emph{type}], where \emph{name} is the name of the source that is being referred to and \emph{type} can be one of the following: \emph{article, book, research, survey, stats, numbers, laws, other}, quoting an expert/authority figure, etc.

\paragraph{Examples}

\begin{compactenum}
    \item \hebgloss{בלי ספק, הנושא הזה של שוויון בתעריפי נסיעה הוא נושא חשוב.}
    {Without a doubt, this issue of fare parity is an important one.}
    \fval{stance confidence level}{high}, 
    \fval{stance type}{epistemic}, 
    \fval{polarity}{(positive,)}.
\end{compactenum}

\subsection{Hedging}
This component highlights lexical items used by speakers to express a low level of commitment towards their claim.

\paragraph{Hedge} The lexical item used for hedging.
A partial list of hedges includes 
\hebgloss{הרבה}{many}, \hebgloss{אחרים}{others}, 
\hebgloss{לעיתים}{often}, 
\hebgloss{לפעמים}{sometimes}, \hebgloss{בערך}{approximately}.

\paragraph{Examples}
\begin{compactenum}
\item \hebgloss{אלה לא בדיוק דמי אבטלה... מענק של עד 4000 שקל}{These are not exactly unemployment benefits; a grant of up to NIS 4,000.}
\fval{hedge}{\hebgloss{עד}{up to}}
\end{compactenum}

\subsection{Quantities}
This component indicates whether the proposition contains quantitative expressions, and if so, how precise they are. The motivation is the assumption that the more accurate a quantity is, the more check worthy the proposition is.
The following features characterize a quantified expression:
\paragraph{Exp} The numerical expression in the sentence. When the expression is a percentage or a fraction, we indicate the whole part it refers to, 
if known, and mark the relation between them. The values of this feature are the literal numeral expressions. 
\paragraph{Quantifier} The quantifier in the sentence. 
Possible values include \hebgloss{כל}{every}, \hebgloss{יש}{there is}, and \hebgloss{מעט}{a few}.

\paragraph{QuantifierType} The type of the quantifier, if one exists. Values are \emph{universal, existential}, and \emph{partial}.

\paragraph{Accuracy} Is the quantity an estimation, an accurate one, or obscured. Values are \emph{accurate, estimate}, and \emph{obscure}. This feature is only valid for numerical quantities, not for quantifiers.

\paragraph{Examples}
\begin{compactenum}
    \item \hebgloss{יש מאות ואלפי אנשים שפרשו לפנסיה, וכעובדה, אין להם פנסיה.}{There are hundreds and thousands of people who have retired, and as a fact, they have no pension.}
    \fval{quantity exp}{\hebgloss{מאות}{hundreds}}, \fval{accuracy}{estimate}, 
    \fval{quantity exp}{\hebgloss{אלפי}{thousands}}, \fval{accuracy}{estimate}.
\end{compactenum}

\subsection{Named entities}
This component highlights named entities and specifies their type. 
As the Knesset Corpus is already annotated for named entities, we simply adopted the annotation, correcting it in rare cases.

\paragraph{Name} The tokens of which the entity consists.
\paragraph{Type} The type of the named entity, one of the types defined by the \href{https://github.com/IAHLT/hebrew_named_entities_open_dataset/blob/main/he-guidelines.pdf}{IAHLT NER guidelines}.



\subsection{Time Expressions}
The time at which the proposition is claimed to have happened (or will happen) helps determine how check worthy it is. 
Time expressions can be exact dates or parts thereof (e.g., \textnl{1994}); relative time expressions like \textnl{last month}, \textnl{next year}, or \textnl{the year the war started}; fuzzy time expressions like \textnl{(sometime) in the past five years}; etc.

Time expressions may describe a time-range, such as \hebgloss{במהלך השנה האחרונה}{during last year}, \textnl{2018-2019}, etc. When a time range is expressed we mark the start point and the end point of the range. If one of them is missing from the data, we leave it empty; e.g., in \hebgloss{בשנים האחרונות}{in recent years} we can assume the end point is the time of utterance, but we do not know for sure what is the start point of this time range.

When possible, we infer date and time information from the meta-data. E.g., expressions such as \hebgloss{אתמול}{yesterday} are interpreted relatively to the actual day of the session in which the text is included.

\paragraph{TimeExp} The date and time expression, in the format \texttt{YY:MM:DD:HH:MM:SS},
where unspecified or missing components of the timeExp are left empty.
\paragraph{Token} The tokens that make up the TimeExp.

\paragraph{Examples}
\begin{compactenum}
    \item \hebgloss{ב-03 בינואר משרד הבריאות הודיע שהגעת וירוס קורונה לישראל היא עניין של זמן.}{On January 30, the Ministry of Health announced that the arrival of the corona virus in Israel is a matter of time.}
    \fval{timeExp token}{\hebgloss{03 בינואר}{30th January}},  \fval{timeExp}{2020:01:30:::}.
\end{compactenum}

Sometimes the speaker refers to the current event at which they are speaking  using time and location expressions such as \hebgloss{כאן/פה}{here} and \hebgloss{עכשיו}{now}. We mark such expressions when we are certain that they refer to the current meeting, as this might help fact checkers verify the given statement.


    

\section{Annotator Agreement Evaluation}
\label{sec:annotator_agreement}
Factuality is a complex notion and in particular, check-worthiness is highly subjective 
\citep{10.1145/3412869}.
To assess agreement across annotators we designated~100 sentences that were annotated by each of the three annotators. 
In some cases, annotators did not agree on the number of claims in the sentence, leading to discrepancies in both the number of claims identified and which claims were tagged, complicating the evaluation of inter-coder agreement. We therefore focus on~65 sentences for which all annotators agreed on the set of claims, which serve as our evaluation set. 
We also measured agreement for a subset of the evaluation set, containing 56 single-claim sentences. 

Another challenge in assessing agreement is the complexity of the scheme, which assigns a large, structured tag to each claim. Due to the high number of features, even a single feature with inconsistent annotation would count as disagreement, naturally resulting in a low agreement score that may not accurately reflect the true level of agreement. To mitigate this, we divided the features into \emph{layers}, each containing between~4 and~10 features, and assessed agreement separately for each layer. 
Additionally, we specifically examined agreement on the primary feature, \emph{check-worthiness score}. 

The results for each evaluation set and each layer are presented in Table~\ref{tbl:agreement}. The scores represent the level of agreement among the three annotators: A score of~3 indicates the percentage of sentences in the test set, where all three annotators agreed on the labels of \emph{all} features in the layer. A score of~2 reflects cases where two out of three annotators reached full agreement and a score of~1 signifies that all annotators differed on \emph{at least one} label.

Evidently, for the strictest agreement score (3), the results are relatively low for some of the layers, such as \emph{Check-worthiness}, \emph{Stance} and \emph{Agency and ESP}, highlighting the difficulty of these tasks. 
However, majority agreement---where we count agreement among two or three annotators---is significantly higher. This is especially notable given the large number of features in each set and the fact that many features involve marking specific string spans within the text, where even a single-character difference would be counted as a disagreement.

\begin{table*}[hbt]
\centering
\begin{tabular}{l|lll|lll}
\multicolumn{1}{r|}{\textbf{Test Set}}
& \multicolumn{3}{c|}{\textbf{Full (65 sents.)}} & \multicolumn{3}{c}{\textbf{Single-claim (56 sents.)}}\\ 
\multicolumn{1}{l|}{\textbf{Layer}} & \multicolumn{1}{c}{3} & \multicolumn{1}{c}{2} & \multicolumn{1}{c|}{1} 
 & \multicolumn{1}{c}{3} & \multicolumn{1}{c}{2} & \multicolumn{1}{c}{1} \\ 
\hline
\emph{check-worthiness score} & 56.92 & 43.08 & 0 & 58.93 & 41.07 &0\\
Check-worthiness & 30.77 & 47.69 & 21.54  & 35.71 & 48.21 & 16.07\\
Agency and ESP & 35.38 & 36.92 & 27.69 & 37.50 & 29.29 & 23.21 \\
Stance & 35.38 & 41.54 & 23.08 & 39.29 & 44.64 & 16.07 \\
Quantities & 67.69 & 32.31 & 0 & 69.64 & 30.36 & 0 \\
Time Expressions & 92.31 & 4.62 & 3.08 & 94.64 & 3.57 & 1.79 \\
\end{tabular}
\caption{Annotator agreement score (\% of sentences annotated identically) by layer and test set. 
}
\label{tbl:agreement}
\end{table*}

We also calculated mean pairwise agreement, using Kappa \citep{doi:10.1177/001316446002000104},
among the annotators and between each annotator and the other two, on the \emph{check-worthiness score} feature, to asses how consistently the annotators agreed on their judgments. 
%
%
%
%
%
%
Mean pairwise Kappa between each of the annotators and the other two ranged between~0.5 and~0.63; overall, among the three annotators, it was~0.58.
The similar agreement scores across annotators 
indicate a consistent annotation process with reasonable reliability.



\section{Annotation of the Knesset Corpus}
\label{sec:experiments}
Our ultimate goal is to automatically annotate all the sentences in the Knesset Corpus with the complex, multifaceted  factuality tag, as  detailed in \S\ref{sec:schema}. This is a large-scale task that will require multiple models; we are initially focusing on our primary feature, \emph{check-worthiness score}. This feature has three possible values (\S\ref{sec:schema}): \emph{worth checking}, \emph{not worth checking} and \emph{not a factual proposition}.

First, we used off-the-shelf GPT models to annotate the corpus for this feature (\S\ref{sec:gpt annotations}). Due to the limited success, we transitioned to more traditional model training approaches(\S\ref{sec:fine-tuned models}), significantly improving the results.

\subsection{Annotation with GPT Models}
\label{sec:gpt annotations}
As an initial approach, we experimented with GPT models to predict the \emph{check-worthiness score} feature. Our goal was to evaluate whether these models could provide accurate predictions, potentially reducing the need for extensive manual labeling. We conducted several experiments using GPT-4 \citep{openai2024gpt4technicalreport} and \href{https://openai.com/index/hello-gpt-4o/}{GPT-4o}, 
which are considered state-of-the-art for similar tasks. We set the temperature of the models to zero in order to ensure deterministic outputs. We evaluated these models on the one-claim sentences evaluation set described in \S\ref{sec:annotator_agreement}, which contains 56 sentences.  We chose this set because it is both reliable, having been annotated by all three annotators, and relatively simple, as each sentence contains only one claim, making it suitable for establishing a baseline for these GPT models. The predictions generated by the models were then compared to the majority vote of the human annotators. We now describe the different experiments we conducted with these models in an attempt to achieve the highest performance. The full results of all the experiments are compiled in Table~\ref{tbl:gpt accuracy}.%
\footnote{The evaluation set is unusually small, but it is high-quality, as these sentences were annotated by three experts. Given that the GPT models perform so poorly on the relatively easy task of annotating a single-claim sentence, there was not much point testing them on a more complex test set.
We do test our fine-tuned models more rigorously (\S\ref{sec:fine-tuned models}).}

\begin{table*}[hbt]
\centering
\begin{tabular}{lrrrrr}
& \multicolumn{2}{c}{\textbf{Accuracy}} & \multicolumn{2}{c}{\textbf{Kappa}} 
\\
\multicolumn{1}{c}{\textbf{Prompting Technique}} & \multicolumn{1}{c}{\textbf{GPT 4}} 
& \multicolumn{1}{c}{\textbf{GPT 4o}}  & \multicolumn{1}{c}{\textbf{GPT 4}} 
& \multicolumn{1}{c}{\textbf{GPT 4o}} \\
Zero-shot & 30.4 & 37.5 & 0.09 & 0.16\\
Hebrew prompts, English labels & 35.7 & 37.5  &0.14 & 0.13\\
Hebrew prompts, Hebrew labels & 28.6 & 37.5 & 0.06 & 0.12\\
Instruction-Based & 32.1 & 48.2 &0.10 & 0.28 \\
Few-shot & 60.7 & 58.9 & 0.25 & 0.25\\
\end{tabular}
\caption{GPT models accuracy, as evaluated on annotators' majority vote, and mean pairwise Kappa agreement between models and annotators. Recall that Kappa for human annotators ranged between~0.5 and~0.63.}
\label{tbl:gpt accuracy}
\end{table*}


\paragraph{Zero-Shot Prompting}
We first established a baseline for the models by evaluating their performance without providing explanations or additional examples. The prompts used for this experiment are listed 
in Appendix~\ref{app:zero-shot prompt}.

Table~\ref{tbl:gpt accuracy} depicts the results: accuracy of~37.5\% 
with GPT-4o and~30.4\% 
with GPT-4.
To better interpret these results, we also calculated the mean pairwise Kappa agreement between the model and each of the annotators. 
This allowed us to assess the annotation quality in comparison to human annotation. The results, presented in Table~\ref{tbl:gpt accuracy}, indicate that the models performed significantly worse compared to human annotators.


\paragraph{Hebrew Prompting}
The sentences we annotate are in Hebrew; we hypothesized that using Hebrew prompts might improve model performance, following \citet{behzad-etal-2024-ask} who suggested that prompting in a different language may yield better results. To test this, we conducted experiments using two different setups: \begin{inparaenum}
    \item A Hebrew prompt with labels remaining in English.
    \item A Hebrew prompt with labels also translated into Hebrew.
\end{inparaenum}
The prompts used in these experiments are similar to those in the zero-shot experiment, but are translated to Hebrew.
The results of these experiments are presented in Table~\ref{tbl:gpt accuracy}. In most cases Hebrew prompts did not improve the results compared to the English prompts, suggesting that the models' performance is not significantly influenced by the prompt language in this case.

\paragraph{Instruction-based Prompting}
We conducted an additional experiment where we provided the model with explicit definitions for each label, along with simple examples, before asking it to classify sentences.

The prompts used in this experiment are given in the Appendix~\ref{app:instruction prompt}.

The accuracy with this approach was~48.2\% 
for GPT-4o and~32.1\% 
for GPT-4, indicating a slight improvement in performance.
A minor improvement is evident also in the mean pairwise agreement scores.

\begin{table*}[htb]
\centering
\begin{tabular}{lrrrrr}
& \multicolumn{2}{c}{\textbf{Accuracy}} & \multicolumn{2}{c}{\textbf{Kappa}} 
\\
\textbf{Model} & \textbf{Test set} & \textbf{Single-claim Set}  & \textbf{Test set} & \textbf{Single-claim Set}\\
AlephBertGimmel & 74.61 & 76.79 & 0.59& 0.61 \\
DictaBERT & 75.50 & 76.79 & 0.60 & 0.62\\
Knesset-DictaBERT & 77.26 & 78.57 & 0.63 & 0.63\\
\end{tabular}
\caption{Accuracy and Kappa agreement of fine-tuned models on the test set and on the single-claim evaluation set, according to annotators' majority vote in the case of the single-claim set, and a single annotation in the case of the larger test set. Recall that Kappa for human annotators ranged between~0.5 and~0.63. 
}
\label{tbl:finetuned accuracy}
\end{table*}

\paragraph{Few-shot Prompting}
In this experiment we also included four real examples from the annotated dataset (disjoint from the evaluation set of course) for each label, in addition to the explanations provided in the previous experiment.  
The prompts given to the models are presented in the Appendix~\ref{app:few-shot prompt}.
This approach led to a significant improvement in results, with GPT-4o achieving 58.9\% accuracy 
and GPT-4 achieving 60.7\% accuracy. 
Kappa scores are~0.25, better than the previous approaches but still far below human agreement.
%
We conducted additional experiments with varying numbers of examples from the dataset, but these variations did not lead to a significant change in results.

\subsection{Experiments with Fine-tuned Models}
\label{sec:fine-tuned models}
Since prompt engineering for GPT models did not yield satisfactory results, we turned to pre-trained encoder-based Hebrew models and fine-tuned them for our task. Specifically, we experimented with AlephBertGimmel \citep{gueta2022large}, DictaBERT \citep{shmidman2023dictabert} and Knesset-DictaBERT  \citep{goldin2024knessetdictaberthebrewlanguagemodel} models. 
To train and evaluate these models, we used the~4,987 sentences that had been manually annotated by human annotators as described in \S\ref{data}, excluding the sentences that were used for annotation agreement evaluation. 

Since the \emph{check-worthiness score} feature is annotated for each claim in a sentence, while we wanted to annotate each sentence with a single value, we adjusted the labels for each sentence as follows: if at least one claim in the sentence was labeled as \emph{worth checking}, the label given to the sentence was \emph{worth checking}. If neither one of the claims is check worthy, but at least one of the claims was labeled as \emph{not a factual proposition}, then this was the label given to the sentence. Otherwise, the label was \emph{not worth checking}.

The models were trained on 80\% of the dataset, while 10\% was used for tuning and the remaining~10\% (488 sentences) constitute the test set. We also evaluate our models on the evaluation set of 56 single-claim sentences (\S\ref{sec:annotator_agreement}). 
The full details of the training process,
are described in Appendix~\ref{app:finetune_training}. 
along with a set of fully‐annotated examples, are available on our \href {https://github.com/HaifaCLG/Factuality}{GitHub repository}.

The classification results for these datasets are presented in Table~\ref{tbl:finetuned accuracy}.
They indicate that fine-tuned models achieved significantly better results compared to GPT models. This suggests that, despite GPT’s proven strengths across many NLP tasks, for this challenging and non-trivial classification task, fine-tuning models on labeled data provides a clear advantage. 
Among the models tested, Knesset-DictaBERT achieved the highest accuracy. This is unsurprising, as this model underwent domain adaptation specifically tailored to Knesset data. Given its superior performance, we will use this model to annotate the entire Knesset Corpus for check-worthiness.



\section{Conclusion}
We presented a complex annotation scheme for factuality and a set of~4,987 manually-annotated sentences, of which~100 are annotated thrice. These resources are open and can be used to train fact-checking applications.
We release the  \href{https://huggingface.co/datasets/GiliGold/Knesset_check_worthiness}{full Knesset Corpus automatically annotated for check-worthiness} and the  \href{https://huggingface.co/GiliGold/knesset-dicta-checkworthiness}{fine-tuned Knesset-DictaBERT model}.
We also release \href {https://github.com/HaifaCLG/Factuality}{a set of annotated examples}
and \href {https://github.com/HaifaCLG/Factuality}{the prompts used to train the models}.

This is an ongoing project, and our current effort centers on developing methods for predicting the complex annotation tags that our scheme defines on a large corpus of parliamentary proceedings texts. The initial experiments that we reported on here are promising, but they are limited to a single feature of the scheme (albeit a critical one), and in future work we would like to predict the full factuality structures.
Other plans for future work include adaptation of our scheme to other languages, with a focus on morphologically-rich languages.
Finally, we plan to explore how factuality is manifested in different groups of the parliament (e.g., government vs.\ opposition).


\paragraph{Limitations}
Our study provides a valuable schema for factuality annotation and preliminary experimental results, yet it has several limitations worth discussing. First, we automatically annotated the Knesset Corpus for only one aspect of the schema,  the check-worthiness score. Future work will focus on developing and refining models for the automatic annotation of other schema elements. Second, our annotation experiments with contemporary LLMs have been limited to GPT-4o. Finally, while we believe the schema could be applied to other domains and languages, this has yet to be demonstrated.


\paragraph{Ethical Considerations}
Our main dataset is the Knesset Corpus which is open and publicly available. Adding annotation of factuality cannot, in our eyes, be abused or dual-used.

We employed three linguists (two women, one man, all native Hebrew speakers residing in Israel) as annotators. 
They were recruited directly (i.e., not via any crowd-sourcing platform) and were paid an hourly wage that is approximately 2.5 times the minimum wage in Israel. No human participants were required for this project.


\section*{Acknowledgments}
We are indebted to Piroska Lendvai for suggesting the topic of this research and for extensive collaboration during the initial phases of the project. Many of the ideas we implemented are hers. 
We are grateful to Israel Landau and Avia Vaknin for their meticulous annotation efforts.
We thank the Idit PhD Fellowship Program at the University of Haifa for supporting the first author. 
This research was supported by the Ministry of Science \& Technology, Israel under grant no.~3-17990.

\bibliographystyle{acl_natbib}

\bibliography{anthology,custom}

\clearpage

\appendix

\section{GPT Classification Prompts}\label{app:prompts}
This section presents the prompts used to classify sentences for check-worthiness using GPT-based models, as desribed in \S\ref{sec:gpt annotations}.

The \emph{User Prompt} used in all experiments is as follows:

\begin{quote}
    \textnl{Classify the following sentence into one of the following categories: `worth checking', `not worth checking', or `not a factual proposition'. Respond with only the label. Sentence: \texttt{\{sentence\_input\}}}
\end{quote}

The \emph{System Prompts} for each prompting technique are described below:

\paragraph{Zero-Shot Prompt}\label{app:zero-shot prompt}

\begin{quote}
    \textnl{You are a helpful assistant that strictly classifies sentences into `worth checking', `not worth checking', or `not a factual proposition'. Always respond with only the label and nothing else.}
\end{quote}


\paragraph{Instruction Based Prompt}\label{app:instruction prompt}

 \begin{quote}
 \textnl{You are a helpful assistant that strictly classifies sentences into `worth checking', `not worth checking', or `not a factual proposition'. Always respond with only the label and nothing else. Follow the definitions provided strictly and always respond with only the label.   
Here is what each category means:
        1. `worth checking' - Sentences that include claims or propositions that can be factually verified or debunked.
        For example: `The Earth is flat.'
        2. `not worth checking' - Sentences that include obvious truths or widely accepted facts, 
        or subjective opinions or claims that cannot be verified or are not important to check. 
        For example: `The sun rises in the east.' or `Chocolate ice cream is the best dessert.'
        3. `not a factual proposition' - Sentences that do not propose a factual claim. 
        This includes questions, commands, or exclamations. 
        For example: `Do you think this is true?' or 'Please close the door.' }
\end{quote}

\paragraph{Few-Shot Prompt}\label{app:few-shot prompt}
These examples were originally in Hebrew, but are presented here in English (translated to English using gpt-4o).

\begin{quote}
    \textnl{
    You are a helpful assistant that strictly classifies sentences into `worth checking', `not worth checking', or `not a factual proposition'. Always respond with only the label and nothing else. Follow the definitions provided strictly and always respond with only the label.   Here is what each category means:
        1. `worth checking' - Sentences that include claims or propositions that can be factually verified or debunked.      
    Examples:  }
    
   `- Additionally, during the Eighteenth Knesset, the Ministry of Justice promoted reforms in various legislative areas: criminal law, security, civil law, economic-fiscal law, administrative law, and international law.'
   
   `- The budget is the best way for a government to express its vision, priorities, and its approach to shaping society.'  
   
   `- Seventy members of my family were murdered for the sanctification of God's name in that horrific place called Auschwitz-Birkenau.'  
   
   `- The State of Israel is now 66 years old, I believe.'  
   
\textnl{    2. `not worth checking' - Sentences that include obvious truths or widely accepted facts, or subjective opinions or claims that cannot be verified or are not important to check. Examples: }
    
    `- I am saying here that we have a shared responsibility with the Ministry of Finance to advance this.'
    
    `   - As the Chairperson of the Public Petitions Committee, you surely know how burdensome it is for the public when they feel they are paying different amounts for the same service.'
    
     `-I don’t even want to imagine.'
        
    `   - He is afraid of everything.'
    
\textnl{3. `not a factual proposition' - Sentences that do not propose a factual claim. This includes questions, commands, or exclamations. Examples: }
    
    `- To you, members of the Nineteenth Knesset, I will recommend and request: focus on significant legislative proposals, substantive issues, and comprehensive, important reforms.  
    
   `- Where is the Minister of Finance to come up and respond?'  
   
   `- Thank you.'
   
   `- So do me a favor, stop with this whole `he's a handsome guy, a nice guy,' and that he provided cellular security for people in Israel.'  
\end{quote}

\section{Fine-tuning Training Process}\label{app:finetune_training}
We fine-tuned three transformer-based models: dicta-il/dictabert \citep{shmidman2023dictabert}, GiliGold/Knesset-DictaBERT \citep{goldin2024knessetdictaberthebrewlanguagemodel}, and dicta-il/alephbertgimmel-base \citep{gueta2022large}. All models were fine-tuned under the same conditions using the Hugging Face Transformers library \citep{wolf-etal-2020-transformers}. We used the manually annotated sample of the Knesset corpus, as described in \S\ref{data} as our dataset. The dataset was split into 80\% training, 10\% validation sets and 10\% test set. Each sentence was tokenized with a maximum sequence length of 512 tokens.
Training was conducted for 3 epochs with a batch size of 8 and a weight decay of 0.01. We used the \href{https://huggingface.co/docs/transformers/main_classes/trainer}{Hugging Face Trainer API} with evaluation and model checkpointing performed at the end of each epoch. The best model was selected based on validation accuracy.
\section{More detailed definitions}

\subsection{Claim type}
\label{app:konst}
\citet{10.1145/3412869} defined seven types of claims:
\begin{description}
\item[Personal experience] Claims that are not capable of being checked using publicly-available information,
e.g., ``I can’t save for a deposit.''
\item[Quantity in the past or present] Current value of something, e.g., ``1 in 4 wait longer than 6 weeks to be seen by a doctor.'' Changing quantity, e.g., ``The Coalition Government has created 1,000 jobs for every day it’s been in office.'' Comparison, e.g., ``Free schools are outperforming state schools.'' Ranking, e.g., ``The UK’s the largest importer from the Eurozone.''
\item[Correlation or causation] Correlation, e.g., ``GCSEs are a better predictor than AS if a student will get a good degree.'' Causation, e.g., ``Tetanus vaccine causes infertility.'' Absence of a link, e.g., ``Grammar schools don’t aid social mobility.''
\item[Current laws or rules of operation] Declarative sentences, which generally include the word ``must'' or
legal terms, e.g., ``The UK allows a single adult to care for fewer children than other European countries.''
Procedures of public institutions, e.g., ``Local decisions about commissioning services are now taken by
organisations that are led by clinicians.'' Rules and changes, e.g., ``EU residents cannot claim Jobseeker’s
Allowance if they have been in the country for 6 months and have not been able to find work.''
\item[Prediction] Hypothetical claims about the future, e.g., ``Indeed, the IFS says that school funding will have fallen by 5\% in real terms by 2019 as a result of government policies.''
\item[Other type of claim] Voting records, e.g ``You voted to leave, didn’t you?'' Public opinion, e.g ``Public
satisfaction with the NHS in Wales is lower than it is in England.'' Support, e.g., ``The party promised free childcare'' Definitions, e.g., ``Illegal killing of people is what’s known as murder.'' Any other sentence that you think is a claim.
\item[Not a claim] These are sentences that do not fall into any categories and are not claims. E.g., ``What do you think?.'', ``Questions to the Prime Minister!''
\end{description}
\section{Additional Resources}
\subsection{SIP List}
We release a list of 313 source introducing predicates (SIPs) \citep{sauri:2009} in English and Hebrew, which we collected (\href {https://github.com/HaifaCLG/Factuality}{GitHub repository}). This list is based on \citet{sauri:2009}'s published \emph{fb\_sip.txt} file, which consists of sentences from FactBank \citep{sauri:2009}, where each event is annotated for SIPs. We extracted all SIPs from this file, removed duplicates, sorted them alphabetically, and manually translated them into Hebrew. 

\subsection{Effective and Epistemic Stance Markers List}
We release a list of predicates, both in the original English and in Hebrew translation, divided into two stance categories: effective stance compatible and epistemic stance compatible. The list was taken directly from \citep{marin2011effective}, and we manually translated it into Hebrew.

\section{Full Examples}\label{app:examples}

A link to the complete set of fully‐annotated examples is available at at \href {https://github.com/HaifaCLG/Factuality}{GitHub repository}.

We list below a few fully annotated examples. These examples also include metadata information, such as the speaker's name or the date of the session, added during a post-processing step. The metadata was not available to the annotators during the annotation process,
ensuring that their judgments remained unbiased,
but it can be incorporated later to enrich the annotation. 

\paragraph{Examples}
\begin{enumerate}

\item \hebgloss{במסגרת רפורמות אלו נקבע תעריף אזורי אחיד עם מעברים חופשיים במשך 09 דקות, כל זאת באזורים המצוינים כאמור, חיפה, גוש-דן וירושלים.}{As part of these reforms, a uniform regional rate was established with free passes for 90 minutes, all in the aforementioned areas, Haifa, Gush Dan and Jerusalem.}
\\check worthiness score = worth checking
\\claim type = Current laws or rules of operation
\\factuality profile = (\hebgloss{סגנית שר התחבורה והבטיחות בדרכים ציפי חוטובלי}{Deputy Minister of Transport and Road Safety Tzipi Hotobli}, CT+)
\\agency agent-less = passive w/o by-clause
\\stance confidence level = high, type = epistemic, polarity = (positive,) 
\\quantity exp = 90, accuracy = accurate
\\entity name = \hebgloss{חיפה}{Haifa}, type = GPE
\\entity name = \hebgloss{גוש דן}{Gush Dan}, type = GPE
\\entity name = \hebgloss{ירושלים}{Jerusalem}, type = GPE

\item \hebgloss{נקבעו ארבע ועדות משנה ואנחנו אמורים להגיש את ההמלצות עד סוף ינואר.}{Four subcommittees have been appointed and we are supposed to submit the recommendations by the end of January.}
\\check worthiness score = worth checking
\\claim type = Quantity in the past or present
\\factuality profile = (\hebgloss{היו"ר דוד צור}{Chairman David Tzur}, CT+)
\\agency agentless = passive w/o by-clause
\\stance confidence level = high, type = epistemic, polarity = (positive)
\\quantity exp = \hebgloss{ארבע}{four}, accuracy = accurate
\\check worthiness score = worth checking
\\claim type = Current laws or rules of operation
\\factuality profile = (\hebgloss{היו"ר דוד צור}{Chairman David Tzur}, CT+)
\\agency agent = \hebgloss{אנחנו}{we}, position = subject, animacy = human, morphology = 1pl
\\stance confidence level = high, type = epistemic, polarity = (positive,)
\\entity name = \hebgloss{ינואר}{January}, type = TIMEX
\\timeExp token = \hebgloss{ינואר}{January}, timeExp = 2014:01:31:::

\item \hebgloss{אני חושב שזה כן נכון וראוי, כמו שאתה אמרת, להשאיר מתוך האנשים האלה שעמדו בכישורים והוועדה איתרה אותם מבלי השפעתו הישירה של השר איזשהו שיקול דעת לתמהיל הנכון.}{I think it is indeed correct and proper, as you said, to leave out of these people who met the qualifications and the committee located them without the direct influence of the minister, any consideration for the right mix.}
\\check worthiness score = not a factual claim
\\claim type = Personal experience
\\factuality profile = (\hebgloss{שר התקשורת גלעד ארדן}{Communications Minister Gilad Erdan}, PR+)
\\agency agent = \hebgloss{אני (גלעד ארדן)}{I (Gilad Erdan)}, position = subject, animacy = human, morphology = 1sg
\\stance confidence level = mid, stance type = effective, polarity = (positive, \hebgloss{כן}{indeed}), reference = (\hebgloss{כמו שאתה אמרת}{as you said}, quoting)

\item \hebgloss{עובדה היא שציבור גדול, אולי ברובו מהציבור החרדי, מעדיף ללכת למוקד של יד שרה, משיקולים שונים – גם רפואיים וגם הלכתיים.}{The fact is that a large public, perhaps mostly from the ultra-orthodox public, prefers to go to the Yad Sara center, for various reasons - both medical and halachic.}
\\check worthiness score = worth checking
\\claim type = Quantity in the past or present
\\factuality profile = (\hebgloss{יואב בן צור (ש"ס)}{Yoav Ben-Tzur (Shas)}, CT+)
\\agency agent-less = nominal sentence
\\stance confidence level = high, stance type = epistemic, polarity = (positive,)
\\hedge = \hebgloss{ציבור גדול}{large public}
\\hedge = \hebgloss{אולי}{perhaps}
\\ entity name = \hebgloss{יד שרה}{Yad Sara}, type = ORG

\item \hebgloss{במאי השנה היו לנו 34 מקרים של התפרצות, וב-7102 במאי לפני שנה היה מקרה אחד.}{In May of this year we had 43 outbreak cases, and in 2017 in May a year ago there was one case.}
\\check worthiness score = worth checking
\\claim type = Quantity in the past or present
\\factuality profile = (\hebgloss{יעל גרמן (יש עתיד) }{Yael German (Yesh Atid)}, CT+)
\\agency agent-less = nominal sentence
\\stance confidence level = high, stance type = epistemic, polarity = (positive,)
\\quantity exp = 43, accuracy = accurate
\\quantity exp = \hebgloss{אחד}{one}, accuracy = accurate
\\entity name = \hebgloss{מאי}{May}, type = TIMEX
\\ timeExp token \hebgloss{מאי השנה}{May of this year}, timeExp = 2018:05::::
\\ timeExp token \hebgloss{7102 במאי}{2017 in May}, timeExp = 2017:05::::

\item \hebgloss{אני רוצה שהוא יגיש היתר בנייה.}{I want him to submit a building permit.}
\\check worthiness score = not worth checking
\\claim type = Other type of claim
\\factuality profile = (\hebgloss{בצלאל סמוטריץ' (הבית היהודי)}{Bezalel Smotrich (The Jewish Home)}, CT+)
\\ESP token = \hebgloss{רוצה}{want}, type = SIP
\\agency agent = \hebgloss{אני (בצלאל סמוטריץ')}{I (Bezalel Smotrich)}, position = subject, animacy = human, morphology = 1sg
\\stance confidence level = high, type = effective, polarity = (positive,)

\item \hebgloss{אני מניח שאתם זוכרים שכאשר כיהנתי כחבר כנסת אחד הנושאים המרכזיים שעסקתי בו זה המאבק בתאונות הדרכים.}{I assume you remember that when I served as a member of the Knesset, one of the main issues I dealt with was the fight against road accidents.}
\\check worthiness score = worth checking
\\claim type = Personal experience
\\factuality profile = (\hebgloss{השר לביטחון הפנים גלעד ארדן}{Minister of Internal Security Gilad Erdan}, PR+)
\\ESP token = \hebgloss{מניח}{assume}, type = SIP
\\ESP token = \hebgloss{זוכרים}{remember}, type = SIP
\\agency agent = \hebgloss{אני (גלעד ארדן)}{I (Gilad Erdan)}, position = subject, animacy = human, morphology = 1sg
\\stance confidence level = mid, stance type = epistemic, polarity = (positive,)
\\hedge = \hebgloss{מניח}{assume}
\\quantity exp = \hebgloss{אחד}{one}, accuracy = accurate

\end{enumerate}
\end{document}